\documentclass{article}
\usepackage{ijcai25}

\usepackage{times}
\usepackage{soul}
\usepackage[hyphens]{url}
\usepackage{xcolor}
\usepackage[hidelinks]{hyperref}
\usepackage[utf8]{inputenc}
\usepackage[small]{caption}
\usepackage{graphicx}
\usepackage{amsmath}
\usepackage{amsthm}
\usepackage{booktabs}
\usepackage{algorithm}
\usepackage{algorithmic}
\usepackage[switch]{lineno}

\title{Integrating Neurosymbolic AI in Advanced Air Mobility: A Comprehensive Survey}

\author{
Kamal Acharya$^1$
\and
Iman Sharifi$^2$\and
Mehul Lad$^1$\and
Liang Sun$^3$\And
Houbing Song$^1$\\
\affiliations
$^1$Department of Information Systems, University of Maryland, Baltimore County\\
$^2$Department of Mechanical and Aerospace Engineering, The George Washington University\\
$^3$Department of Mechanical Engineering, Baylor University\\
\emails
\{kamala2, du72811,songh\}@umbc.edu,
i.sharifi@gwu.edu,
liang\_sun@baylor.edu
}

\begin{document}

\maketitle

\begin{abstract}
Neurosymbolic AI combines neural network adaptability with symbolic reasoning, promising an approach to address the complex regulatory, operational, and safety challenges in Advanced Air Mobility (AAM). This survey reviews its applications across key AAM domains such as demand forecasting, aircraft design, and real-time air traffic management. Our analysis reveals a fragmented research landscape where methodologies, including Neurosymbolic Reinforcement Learning, have shown potential for dynamic optimization but still face hurdles in scalability, robustness, and compliance with aviation standards. We classify current advancements, present relevant case studies, and outline future research directions aimed at integrating these approaches into reliable, transparent AAM systems. By linking advanced AI techniques with AAM’s operational demands, this work provides a concise roadmap for researchers and practitioners developing next-generation air mobility solutions.

\end{abstract}

\section{Introduction}
Artificial Intelligence (AI) has undergone significant development, marked by a "Cold War" between symbolic and neural approaches. The symbolic approach dominated the early stages of AI research, but in recent years, neural models have gained prominence. Both approaches face distinct challenges, and this has paved the way for the emergence of the third wave of AI, known as Neurosymbolic AI \cite{garcez2023neurosymbolic}. Neurosymbolic AI combines the learning capabilities of neural models with the reasoning abilities of symbolic models. Simultaneously, the aviation and transportation industries are experiencing a transformative shift with the advent of Advanced Air Mobility (AAM). AAM envisions a future where air transportation is seamlessly integrated into everyday life, enabling efficient, safe, and sustainable movement of people and goods. This vision is being realized through technological advancements in electric propulsion, automation, and materials science, fostering the development of innovative aircraft such as electric vertical take-off and landing (eVTOL) vehicles \cite{johnson2022nasa}.

AAM faces several obstacles like regulatory complexity, the need for safe and explainable automated decision-making, and the integration of heterogeneous data from demand forecasting, infrastructure constraints, weather, and traffic conditions\cite{cohen2024advanced}. Neurosymbolic AI offers a promising solution by combining the pattern-recognition power of neural networks with the rule-based logic of symbolic systems, allowing for a more interpretable, high-level reasoning framework that can encode safety protocols, regulatory constraints, and domain-specific knowledge. As a result, decisions made by Neurosymbolic AI are not only data-informed but also grounded in explicit safety requirements and operational guidelines critical for fostering growth in AAM by improving trust, compliance, and the overall reliability of autonomous aerial operations.

Neurosymbolic AI offers significant potential for AAM by combining neural learning from large, heterogeneous datasets (e.g., sensor streams, flight trajectories, air traffic patterns) with symbolic reasoning that enforces regulatory and safety constraints. This dual approach supports both Urban Air Mobility (UAM) and Regional Air Mobility (RAM) by ensuring models not only adapt to complex, real-time data but also adhere to airspace rules, vehicle performance limits, and scheduling constraints. In particular, Neurosymbolic Reinforcement Learning \cite{acharya2023neurosymbolic} can optimize flight paths, manage dynamic air traffic, and coordinate ground infrastructure in real time, balancing efficiency with safety and noise reduction. Embedding domain knowledge into the symbolic layer alongside adaptive neural policies allows continuous strategy refinement in AAM systems. However, developing robust Neurosymbolic architectures for challenges like multi-vehicle coordination, real-time constraint satisfaction, and large-scale deployment remains an open research area. This article categorizes key research domains essential for advancing AAM from engineering design and electrification to demand modeling, autonomy, cybersecurity, and public acceptance and outlines how Neurosymbolic AI can be applied across these areas, along with potential case studies and associated risks.

While surveying existing literature, we observed no comprehensive review on the integration of Neurosymbolic AI within AAM. Prior studies have focused on isolated topics, demand modeling \cite{LONG2023102436}, aircraft design \cite{KIESEWETTER2023100949}, electrification \cite{dorn2020power}, and safety \cite{yoo2022risk}, but none have explored how Neurosymbolic AI can accelerate and enhance research across all AAM domains. This survey paper is the first to provide a detailed analysis of ongoing research efforts in AAM and to propose ways Neurosymbolic AI can drive advancements in this field. 

\section{Background}
\subsection{Neurosymbolic AI}
\begin{figure}
    \centering
    \includegraphics[trim=20mm 24mm 15mm 20mm, clip, width=\linewidth]{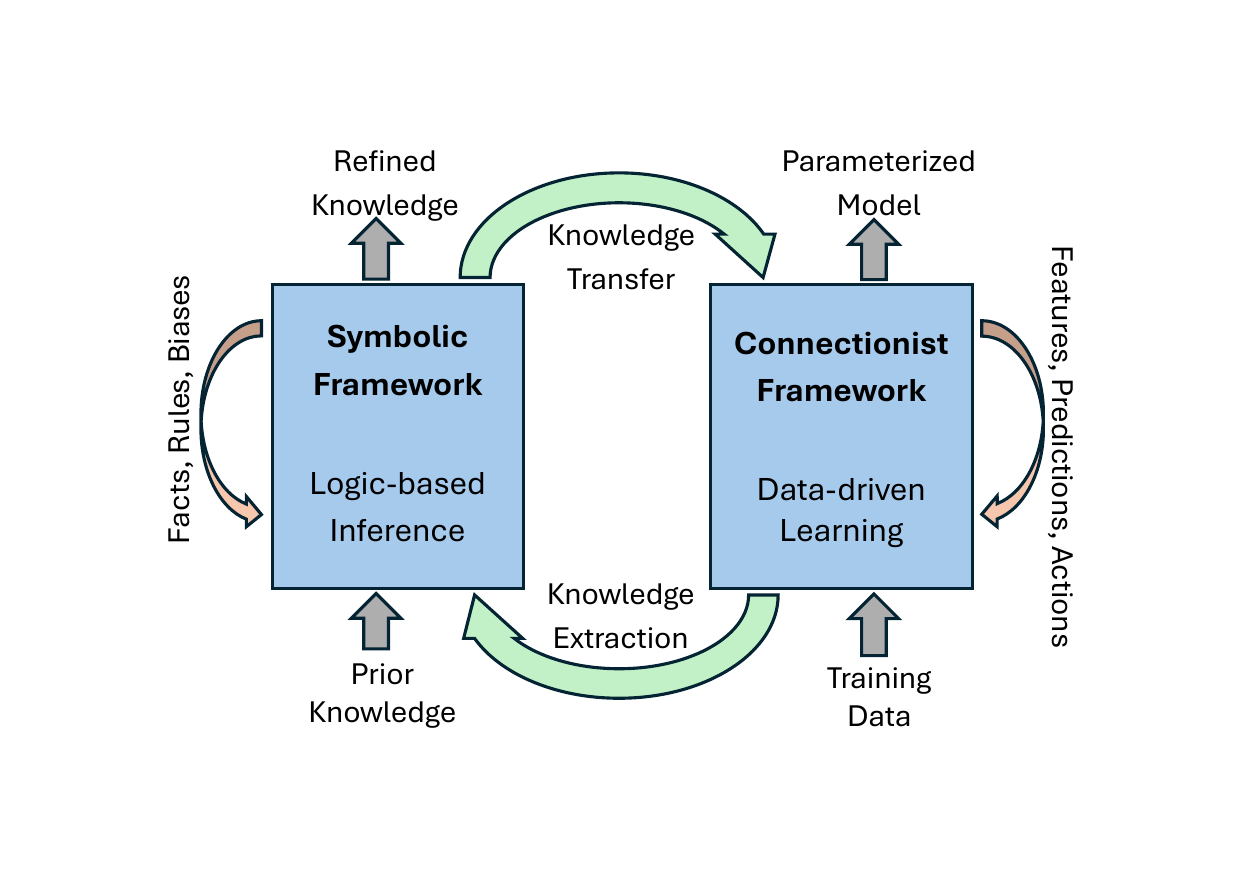}
    \caption{General interactions between symbolic and connectionist frameworks in Neurosymbolic AI}
    \label{nesy}
\end{figure}

Neurosymbolic AI represents a hybrid approach that integrates neural networks' learning capabilities with symbolic AI’s structured reasoning, overcoming the limitations of both paradigms. Symbolic AI, dominant from the 1950s to the 1980s, offers explainable reasoning and structured knowledge representation, but it requires extensive manual rule definitions \cite{mao2019neuro}. In contrast, connectionist AI, or deep learning, enables data-driven pattern recognition but suffers from black-box decision-making and interpretability challenges \cite{acharya2023neurosymbolic}. Neurosymbolic AI leverages the strengths of both by enabling efficient knowledge extraction and conceptual reasoning, requiring less training data while maintaining transparency and adaptability in uncertain environments \cite{garcez2023neurosymbolic}. Due to these advantages, Neurosymbolic AI is often regarded as the third wave of AI, capable of achieving error correction, enhanced explainability, and logical inference beyond deep learning alone. \autoref{nesy} depicts the general interaction between the symbolic and connectionist frameworks in Neurosymbolic AI.

Several taxonomies have been proposed to classify Neurosymbolic AI systems. Henry Kautz proposed a classification system based on six distinctive categories \cite{kautz2022third}. More recently, a survey categorized Neurosymbolic AI based on efficiency, generalization, and interpretability, leading to three primary approaches: learning for reasoning, reasoning for learning, and learning-reasoning \cite{yu2023survey}. Learning for reasoning uses neural networks to enhance symbolic reasoning, reasoning for learning incorporates symbolic logic to refine neural learning processes, and learning-reasoning represents a fully integrated system where both paradigms influence each other dynamically. These frameworks provide a comprehensive approach to structuring Neurosymbolic AI for robust, explainable decision-making in various domains.

Neurosymbolic AI has broad applications, particularly in intelligent transportation systems (ITS) and AAM. In ITS, Neurosymbolic AI enables real-time sensor integration with pre-defined traffic rules, ensuring adaptive yet regulatory-compliant AI-driven transportation solutions \cite{garcez2023neurosymbolic}. This improves traffic condition prediction, autonomous vehicle safety, and public transit optimization. Similarly, in AAM and UAM, Neurosymbolic AI enhances decision-making, safety, and airspace management. It aids Unmanned Aircraft Systems (UAS) in navigating complex environments by incorporating expert-defined flight safety rules and predictive modeling \cite{gilpin2021neuro}. By fusing symbolic reasoning with deep learning, Neurosymbolic AI improves reliability, demand forecasting, and autonomous decision-making in emerging aviation technologies, ensuring intelligent, safe, and interpretable air mobility operations \cite{kohaut2024probabilistic}.

\subsection{Advanced Air Mobility}
\begin{figure}[hbt!]
\centering
\includegraphics[width=\columnwidth]{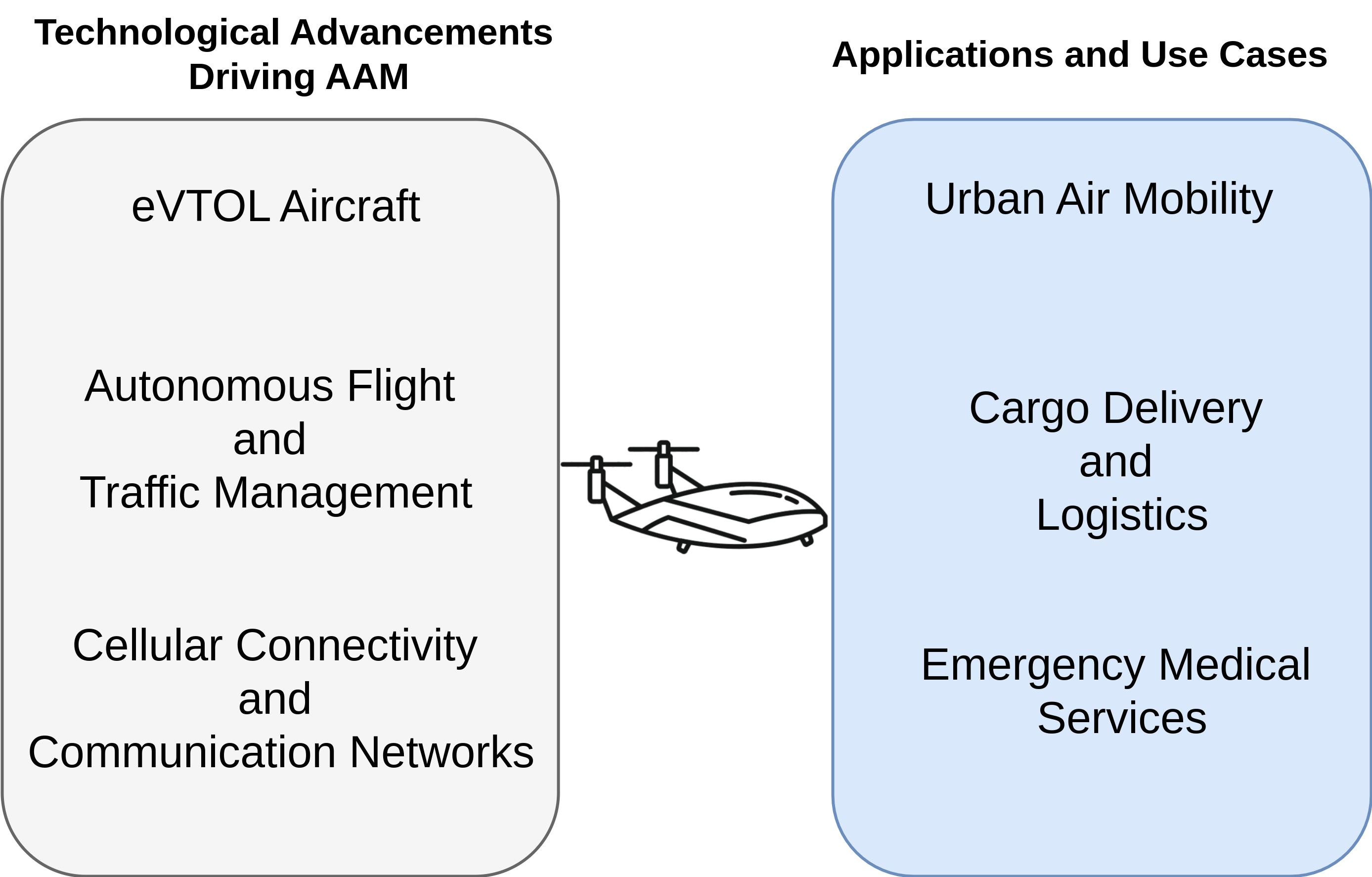}
\caption{AAM Overview}
\label{fig:AAMDevelopment}
\end{figure}

AAM represents a paradigm shift in transportation, promising to revolutionize the movement of people and goods, particularly in urban and regional environments. This transformative vision encompasses a diverse array of applications, extending beyond the initial concept of air taxis to include UAM, cargo delivery, emergency medical services, and various other aerial missions \cite{goyal2022advanced}. \autoref{fig:AAMDevelopment} provides the overview of technologies in advancement of AAM along with the applications areas.

\subsubsection{Technological Advancements Driving AAM}

\paragraph{Electric Vertical Take-Off and Landing Aircraft} eVTOL aircraft are at the heart of the AAM revolution. Their electric propulsion systems offer several compelling advantages over traditional helicopters and airplanes. These include reduced noise pollution, significantly lower emissions contributing to a more sustainable transportation system, and the potential for autonomous operation \cite{LONG2023102436}. The reduced noise footprint is particularly important for urban environments, where noise pollution is a major concern \cite{rizzi2022urban}. However, substantial challenges remain in developing batteries with sufficient energy density and range to support commercially viable flight times and distances. Ensuring the safety and reliability of battery technology, including addressing potential failure modes and mitigating risks, is also crucial. Extensive research is underway to address these technological hurdles, focusing on improvements in battery technology, electric motor design, and overall aircraft design optimization \cite{barrera2022next}.

\paragraph{Autonomous Flight and Traffic Management} The safe and reliable integration of numerous autonomous aircraft into existing airspace demands the development of sophisticated Unmanned Traffic Management (UTM) systems \cite{namuduri2023digital}. These systems are crucial for managing air traffic flow, preventing mid-air collisions, and ensuring compliance with safety regulations. Research is actively exploring various approaches to UTM, including centralized and decentralized architectures. Centralized systems offer a more controlled approach to managing air traffic, while decentralized systems offer greater resilience and scalability. The choice between these architectures involves trade-offs between control and resilience, and the optimal approach may vary depending on the specific operational context \cite{murcca2024characterizing}. The integration of machine learning techniques is being explored to improve the efficiency and adaptability of UTM systems \cite{deniz2024reinforcement}. Furthermore, the development of advanced sensor technologies, such as lidar and radar, is crucial for enhancing the situational awareness of autonomous aircraft \cite{karampinis2024ensuring}.

\paragraph{Cellular Connectivity and Communication Networks} Seamless data exchange between aircraft, ground control, and other stakeholders is essential for real-time monitoring, traffic management, and overall system safety \cite{do2024cellular}. The integration of cellular networks and other communication technologies plays a pivotal role in achieving this. The evolution of cellular technology, with 5G and the anticipated arrival of 6G networks, is expected to significantly enhance the capabilities of AAM communication systems. The use of advanced beam forming techniques can improve the efficiency and reliability of cellular connectivity in AAM scenarios. Researchers are also exploring the use of other communication technologies, such as satellite communication, to augment cellular networks and provide broader coverage \cite{do2024cellular}. Vehicle-to-vehicle (V2V) communication is essential for enabling safe and efficient coordination between aircraft \cite{gur2024v2v}.

\subsubsection{Applications and Use Cases}

\paragraph{Urban Air Mobility } UAM aims to alleviate traffic congestion, reduce commute times, and improve accessibility, particularly for areas with limited ground transportation infrastructure \cite{goyal2021advanced}. However, challenges remain in integrating UAM into existing urban environments, managing air traffic density, and addressing public concerns about noise and safety. UAM systems require the development of sophisticated airspace management strategies to handle the high density of aircraft operations. This includes defining flight corridors, establishing separation standards, and managing potential conflicts with other air traffic \cite{murcca2024characterizing}. The development of robust communication and navigation systems is also crucial for ensuring the safety and efficiency of UAM operations \cite{do2024cellular}. UAM operations need to be integrated with ground transportation systems to provide seamless multimodal travel options. This involves the development of vertiports and other infrastructure to facilitate the transfer between air and ground transportation. The integration of UAM into existing urban environments requires careful consideration of land use planning, community engagement, and public acceptance.

\paragraph{Cargo Delivery and Logistics}

AAM offers significant potential for improving cargo delivery and logistics, particularly in remote or underserved areas. Drones and other AAM vehicles can transport goods more efficiently and cost-effectively than traditional ground transportation, especially in areas with challenging terrain or limited road infrastructure. Applications include medical supplies delivery, package delivery, and the transportation of time-sensitive goods \cite{dulia2021benefits}. The integration of AAM into existing logistics networks requires coordination with ground transportation and warehousing systems \cite{raghunatha2023addressing}. The use of autonomous drones for cargo delivery can reduce labor costs and improve efficiency. However, ensuring the safety and security of autonomous drone deliveries is a critical concern \cite{gur2024v2v}.

\paragraph{Emergency Medical Services (EMS)}

eVTOL aircraft and drones can provide rapid transport of patients to hospitals, reducing response times and potentially improving patient outcomes. The use of AAM in EMS also has the potential to improve access to healthcare in underserved communities \cite{bridgelall2024transforming}. AAM based EMS systems require the integration of AAM vehicles with existing EMS infrastructure and protocols. The economic feasibility of AAM based EMS depends on factors such as operating costs, patient transport times, and the potential for improved patient outcomes. The strategic placement of vertiports near hospitals and other healthcare facilities can optimize response times and improve access to care. The development of efficient routing algorithms and real-time traffic management systems is crucial for optimizing AAM-based EMS operations \cite{bridgelall2024transforming}.

\section{Application Areas of Neurosymbolic AI in Advanced Air Mobility }
\begin{figure}[hbt!]
\centering
\includegraphics[width=\columnwidth]{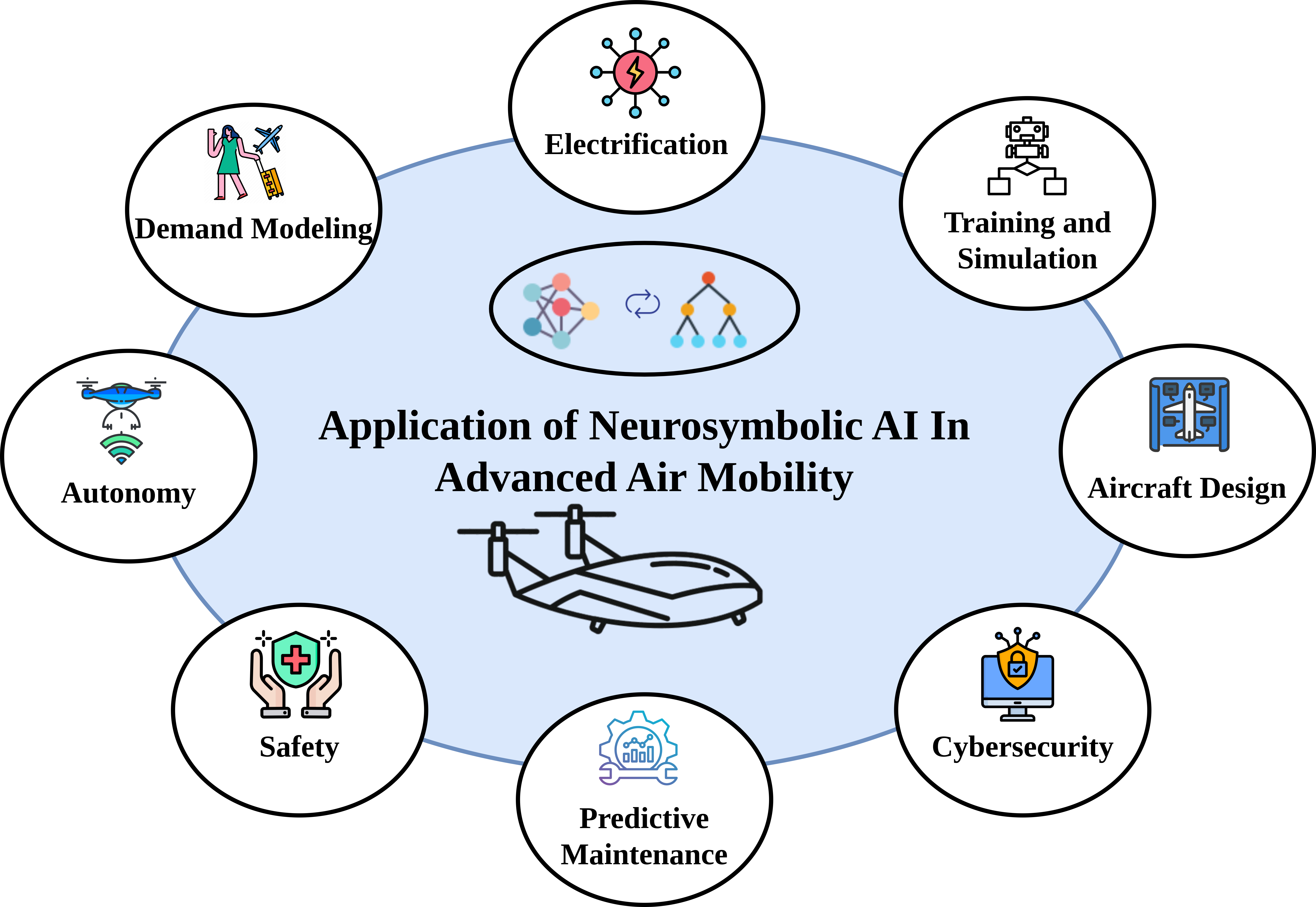}
\caption{Applications of Neurosymbolic AI in AAM}
\label{fig:applications}
\end{figure}

Neurosymbolic AI has been gaining popularity worldwide due to its efficient learning and reasoning capability in almost every domain. In context of AAM, main applications of Neurosymbolic AI is shown in the \autoref{fig:applications}.

\subsection{Electrification}
Electrifying AAM is pivotal for achieving sustainable aviation by reducing emissions, lowering operational costs, and enabling quieter operations. However, the success of eVTOL aircraft is constrained by significant technological and operational challenges, notably the limitations of current lithium-ion battery (LIB) technology. Despite LIBs’ high energy density (300–400 Wh/kg for short-range applications and over 800 Wh/kg for narrow-body aircraft), issues such as weight constraints, energy inefficiency, thermal instability, and inadequate charging infrastructure persist \cite{barrera2022next}. These challenges are exacerbated during high-demand phases like takeoff and landing, necessitating advanced energy management strategies. To address these issues, innovations such as wide bandgap semiconductors, improved thermal management systems, and advanced control strategies are being pursued to enhance the reliability and efficiency of electric propulsion systems \cite{dorn2020power}. Furthermore, Neurosymbolic AI is emerging as a promising tool for optimizing energy usage, extending flight range, and managing battery lifecycles through techniques like predictive energy optimization, smart grid integration, and fault detection.

\subsection{Aircraft Design}
The design of eVTOL aircraft for air taxis and metro systems hinges on advancing battery storage, propulsion efficiency, and noise reduction to meet both safety and urban sustainability criteria. While manufacturers like Airbus are developing specialized eVTOL platforms, challenges persist in certification, infrastructure integration, and public acceptance \cite{silva2018vtol}. Regulatory frameworks need to evolve alongside these technologies, as highlighted by NASA/Deloitte Concept of Operations (CONOPS) \cite{hill2020uam}, and user-centric design elements, such as cabin noise and ride comfort, are critical for adoption. Design trade-offs between payload capacity, flight range, and vehicle weight continue to constrain the scalability of eVTOL systems \cite{KIESEWETTER2023100949}. Moreover, distinct requirements for intra versus inter-city passenger transport, along with the differing design challenges of UAM aircraft and last-mile delivery drones, call for a multifaceted engineering strategy that incorporates aerodynamic efficiency, AI-driven control systems, and optimized energy management. Despite promising advances in control mechanisms, propulsion, and VTOL transition efficiency, bridging the gap between technological potential and real-world deployment remains a significant challenge, underscoring the need for an interdisciplinary approach integrating engineering, policy, and urban planning.

\subsection{Training and Simulation }
Training and simulation are essential for advancing AAM by creating adaptive environments for skill development and operational planning. Neurosymbolic AI, which synergizes the robust pattern recognition capabilities of deep learning with the interpretability and reasoning strengths of symbolic AI, can significantly enhance simulation platforms\cite{siyaev2023interaction}. By integrating Neurosymbolic AI, simulation systems can not only process vast amounts of sensor and operational data through deep learning but also incorporate domain-specific rules and logical reasoning to better replicate the intricate dynamics of AAM environments. This approach enables the development of training scenarios that are both realistic and adaptable, addressing complex operational variables such as weather variations, urban infrastructure constraints, and emergency protocols. Moreover, Neurosymbolic AI facilitates the creation of digital twins that can reason about potential system failures, optimize route planning, and simulate multi-agent interactions with high fidelity\cite{siyaev2023interaction}.

\subsection{Predictive Maintenance}
Predictive maintenance is crucial for enhancing safety and operational efficiency in AAM. By leveraging vast arrays of sensor data and advanced analytics, predictive maintenance systems can identify early indicators of potential component degradation or system malfunctions before they escalate into critical failures\cite{nguyen2022artificial}. Neurosymbolic AI is used in fault diagnosis by integrating a case-based reasoning (CBR) method with a Bayesian network (BN) approach, where fault cases are initially formalized for similarity-based diagnosis and later transformed into a probabilistic model for dynamic multi-fault detection under uncertainty \cite{yang2018optimized}. Neurosymbolic reasoning \cite{siyaev2023interaction} enhances digital twins by enabling natural language interaction, allowing users to manipulate 3D components, read maintenance manuals, and perform installation and removal procedures autonomously.  The result is not only a reduction in unexpected downtime, but also a significant improvement in the lifecycle management of critical systems, ultimately leading to safer and more efficient air mobility operations.

\subsection{Safety }
Safety is a critical aspect of AAM due to the inherent risks and complexities associated with integrating new aerial systems into urban and regional airspaces \cite{yoo2022risk}. According to NASA’s In-Time System-wide Safety Assurance (ISSA) framework, hazards in AAM can arise from multiple sources, including vehicle malfunctions, environmental factors, operational contexts, and broader aviation system challenges \cite{ellis2020time}. The National Transportation Safety Board (NTSB) emphasizes that AAM safety standards must exceed those of general aviation, given the high volume of short-distance flights expected in AAM operations \cite{national2018time}. To ensure safety in AAM, one approach involves using symbolic logic constraints driven by common rules, often derived from Federal Aviation Administration (FAA) regulations, to filter or "shield" the outputs of neural networks, preventing unsafe actions during the training phase \cite{jansen2020safe}. To this end, a Neurosymbolic Deep Reinforcement Learning approach has been proposed to enhance safety by embedding symbolic logic constraints directly into the learning process \cite{sharifi2023towards}. Another strategy involves guiding neural networks using logical safety rules, providing feedback to the AI agent regarding the safety of its actions \cite{kimura2021reinforcement}.

\subsection{Autonomy }
AAM envisions high-density, low-altitude operations in urban and regional environments, requiring aerial vehicles to navigate dynamic obstacles, weather conditions, and air traffic without direct human intervention. Autonomous systems reduce pilot workload, lower operational costs, and enhance safety by making real-time decisions based on sensor data and predefined regulations. Autonomy ensures seamless integration with existing Air Traffic Management (ATM) systems, facilitating coordination between manned and unmanned aerial vehicles (UAVs). By embedding symbolic representations of air traffic rules, safety protocols, and operational guidelines, Neurosymbolic systems can ensure compliance with aviation standards, avoiding collisions and potential conflicts, while dynamically adapting to changing conditions. Moreover, Neurosymbolic AI tackles the overlying scalability issue in traditional connectionist frameworks by integrating symbolic knowledge (e.g., air traffic rules, safety protocols) with neural network-based perception and control systems, allowing autonomous AAM vehicles to scale across diverse environments with minimal retraining \cite{wang2024towards}.

\subsection{Cybersecurity}
AAM cybersecurity challenges stem from their reliance on interconnected systems, autonomous decision-making, and real-time data exchange. Critical threats include GPS spoofing, wireless attacks (e.g., jamming and man-in-the-middle), AI adversarial manipulations, and cyber-physical intrusions, all of which can compromise navigational accuracy and flight control \cite{petit2014potential}. Unlike conventional deep learning models, which are prone to adversarial vulnerabilities, Neurosymbolic AI enhances security by integrating logical rules, knowledge graphs, and domain constraints. For instance, Logic Tensor Networks can merge formal security policies with real-time anomaly detection to filter malicious network activity \cite{grov2024use}, while Neurosymbolic methods also enable cross-validation of multiple navigational inputs to detect and correct GPS spoofing. Neurosymbolic AI driven Security Operations Centers have proven effective in cyber threat intelligence and adaptive incident response, offering a robust defense framework for AAM’s digital infrastructure.

\subsection{Demand Modeling}
AAM demand modeling is challenged by complex, multifaceted data and dynamic operational environments that traditional models often fail to capture \cite{LONG2023102436}. Neurosymbolic AI, which fuses deep neural networks' pattern recognition with the structured reasoning of symbolic systems, has emerged as a promising yet still exploratory approach in this domain \cite{garcez2023neurosymbolic}. Key hurdles include aligning unstructured sensor and user behavior data with the symbolic representations of regulatory and operational constraints, thereby affecting model interpretability, scalability, and robustness. Recent advancements, such as the integration of decision trees with neural networks \cite{acharya2025neurosymbolic} and the development of Probabilistic Mission Design architectures \cite{kohaut2024probabilistic}, have demonstrated improved forecasting accuracy and novel methods for embedding legal frameworks into AI models.

\section{Potential Case Studies}
\subsection{FAA Roadmap for Artificial Intelligence Safety
Assurance}
The FAA Roadmap for AI Safety Assurance\footnote{\url{https://www.faa.gov/aircraft/air_cert/step/roadmap_for_AI_safety_assurance}} establishes a structured, phased approach for integrating artificial intelligence into aviation, a domain traditionally governed by deterministic safety protocols. The roadmap advocates for an incremental, risk-based strategy, initially focusing on low-criticality systems to gather operational data and refine safety assurance processes before transitioning to high-stakes, safety-critical applications. This iterative engineering approach ensures that early-stage insights inform the deployment of more complex AI-driven systems. A key aspect of the roadmap is its distinction between learned AI (static, pre-trained models) and learning AI (adaptive, real-time evolving systems). While learning AI enhances adaptability and responsiveness, it poses significant certification challenges due to its inherent unpredictability, limited explainability, and potential for bias issues that conventional safety methodologies struggle to address.

The roadmap underscores the importance of regulatory alignment with existing aviation safety frameworks, advocating for industry collaboration among regulatory bodies, government agencies, and stakeholders to establish standardized AI assurance protocols. However, it highlights critical gaps, including the lack of robust certification methods for evolving AI, inadequate bias mitigation strategies, and the absence of comprehensive regulatory guidelines for autonomous AI-driven aviation systems. Ethical considerations, essential for public trust and system accountability, receive minimal focus, indicating an area requiring further exploration. Addressing these challenges is essential for ensuring safe, reliable AI integration into aviation while maintaining compliance with rigorous industry standards. By integrating Neurosymbolic AI, the FAA can establish a robust, certifiable AI framework that enhances aviation safety while accelerating the adoption of AI in autonomous flight operations.

\subsection{EASA AI Roadmap}
EASA AI Roadmap 1.0\footnote{\url{https://www.easa.europa.eu/en/downloads/109668/en}} is the initial roadmap established a foundational strategy for integrating AI—primarily machine learning and deep learning—into aviation, addressing its transformative potential across aircraft design, operations, maintenance, air traffic management, drone applications, and safety risk management. It underscored the critical need for robust trustworthiness frameworks through enhanced explainability, human-AI collaboration, and phased certification processes to ensure safety and reliability. EASA AI Roadmap 2.0\footnote{\url{https://www.easa.europa.eu/en/downloads/137919/en}}, expands the scope by incorporating hybrid AI approaches such as Neurosymbolic methods, while further emphasizing cybersecurity and AI safety assurance within a human-centric framework. This version reaffirms AI’s role as an augmentation tool, integrating advanced digital twin technologies and predictive maintenance to enhance operational efficiency and safety. Aligned with the regulatory requirements of the EU AI Act, it introduces risk-based categorization of AI systems, mandating transparency, explainability, and human oversight.

\section{Risk and Challenges}
\begin{figure}[hbt!]
\centering
\includegraphics[width=\columnwidth]{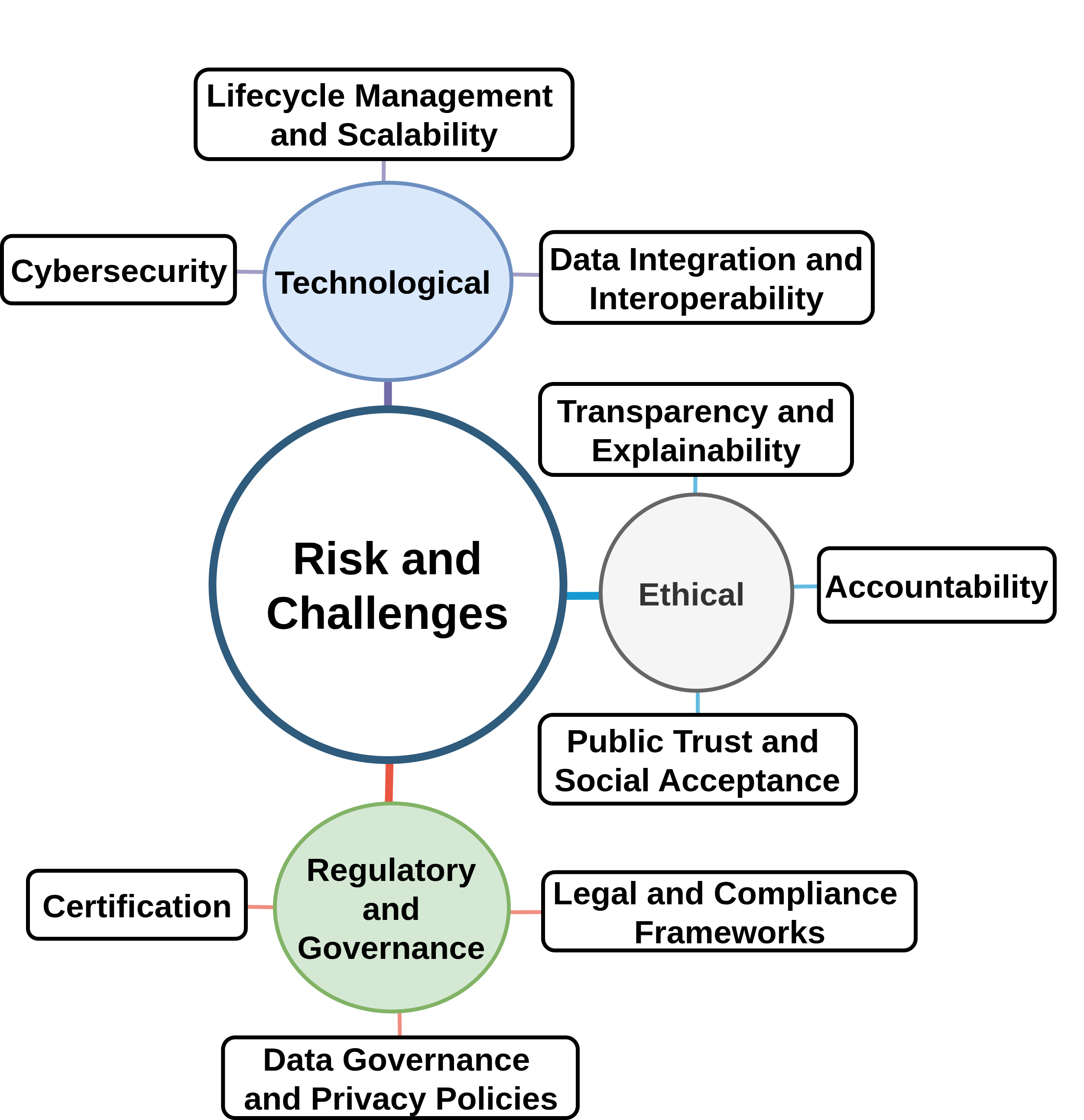}
\caption{Risks and Challenges}
\label{fig:risks}
\end{figure}

Neurosymbolic AI has the potential to enhance the efficiency, safety, and scalability of AAM, but it also presents several challenges and risks as depicted in the \autoref{fig:risks}.

\subsection{Technological Challenges}
\subsubsection{Data Integration and Interoperability}
Data integration in aviation face challenges due to heterogeneous sensor sources. Multi-sensor fusion can be homogeneous or heterogeneous, with the latter requiring precise synchronization. AI enhanced integration with air traffic management demands legacy system re-engineering while ensuring safety compliance\cite{su16198336}. Interoperability is critical, necessitating standardized data formats, exchange protocols, and interface standards. The FAA's Next Gen program focuses on digital system interoperability to support AI-driven airspace operations \footnote{\url{https://www.faa.gov/aircraft/air_cert/step/roadmap_for_AI_safety_assurance}}. Advanced fusion methods leverage variance and covariance estimations for optimal integration, while AI driven standardization enhances cross-platform data accessibility, as seen in the U.S. military’s Combined Joint All Domain Command and Control (CJADC2)  \footnote{\url{https://www.ai.mil/Initiatives/CJADC2/}} system.

\subsubsection{Cybersecurity}
The integration of neural networks and symbolic reasoning, although potentially powerful, introduces new complexities in system design and validation\cite{jalaian_neurosymbolic_2023}. Ensuring the robustness of these hybrid systems against adversarial attacks targeting both the neural and symbolic components requires extensive testing under diverse, real-world conditions \cite{hagos2024neuro}. Moreover, the interpretability of Neurosymbolic models, crucial for safety-critical applications like AAM, remains an ongoing research challenge. The dynamic nature of cyber threats also necessitates continuous adaptation of these systems, raising questions about long-term reliability and maintenance \cite{hagos2024neuro}. The increased computational requirements of Neurosymbolic AI may pose challenges for real-time decision-making in resource-constrained AAM environments\cite{jalaian_neurosymbolic_2023}.

\subsubsection{Lifecycle Management and Scalability}
 As AAM systems continually evolve, the necessity for robust version control, formal verification, and comprehensive testing becomes paramount to maintain both operational integrity and certification. Moreover, the scalability of these frameworks is tested by the heterogeneous nature of AAM fleets, which encompass vehicles with diverse sensor modalities and performance profiles, potentially leading to compatibility issues and performance bottlenecks if not managed through modular and adaptive architectures \cite{kohaut2024probabilistic}. While the modularity of Neurosymbolic systems offers a pathway to seamless integration and system evolution, their practical implementation in AAM must address the inherent trade-offs between adaptability and the rigorous demands of safety and regulatory compliance.

\subsection{Ethical Challenges}
\subsubsection{Accountability}
Accountability issues remain particularly complex in scenarios where autonomous decisions lead to system failures or incidents, determining responsibility among manufacturers, programmers, and operators is nontrivial, highlighting the urgent need for clear regulatory frameworks and transparent decision-making processes that can apportion accountability effectively \cite{novelli2024accountability}. The integration of neural and symbolic methods, while promising enhanced interpretability, does not inherently mitigate risks related to bias and fairness \cite{ferrer2021bias}. Inadequate or skewed training data may propagate discriminatory outcomes, thereby compromising equitable service delivery across diverse operational contexts. Addressing these concerns requires embedding explicit fairness constraints and ethical auditing protocols within the Neurosymbolic framework to ensure decisions are both transparent and impartial.

\subsubsection{Transparency and Explainability for Trust}
Transparency and explainability foster trust and enable scrutiny of automated decision-making processes. By integrating symbolic reasoning with neural networks, these systems can provide more interpretable decision pathways, allowing stakeholders to understand the rationale behind actions taken. Furthermore, implementing robust auditability mechanisms through comprehensive logging and traceability is essential for post-incident reviews, system improvement, and regulatory compliance. These features are vital for maintaining public trust and developing a resilient framework that can adapt to the evolving demands of AAM environments. However, it is important to note that achieving a balance between transparency and system performance remains a challenge, as increasing explainability may impact the efficiency of complex models \cite{acharya2023neurosymbolic}.

\subsubsection{Public Trust and Social Acceptance}
Public trust and social acceptance are crucial for addressing concerns related to data privacy, ethics, and system accountability in the rapidly evolving field of AAM.Despite hybrid approaches like Probabilistic Mission Design (ProMis) \cite{kohaut2024probabilistic} aiming to enhance transparency by merging symbolic reasoning with neural networks, significant risks remain. Current systems still fall short of fully explaining complex decision-making processes, undermining accountability and eroding public trust. Moreover, without sustained, effective collaboration among regulators, manufacturers, and the public, the ethical and responsible deployment of AAM systems is jeopardized, potentially stalling their broader acceptance and market integration. The transparency is particularly important when dealing with sensitive operational data and ensuring compliance with data protection regulations\footnote{\url{https://rm.coe.int/prems-107320-gbr-2018-compli-cahai-couv-texte-a4-bat-web/1680a0c17a}}.

\subsection{Regulatory and Governance Challenges}

\subsubsection{Certification}
The FAA has made limited progress in determining appropriate certification paths for AAM aircraft, partly due to the lack of established airworthiness standards and operational regulations for novel features \footnote{\url{https://www.oig.dot.gov/sites/default/files/FAA_AAM_Certification_Final_Audit_Report_w_NGO_response.pdf}}. This regulatory uncertainty is compounded by the need for new approaches to verify and validate Neurosymbolic systems, which combine probabilistic neural components with deterministic symbolic rules. The development of robust testing protocols that can rigorously verify system behavior under rare or extreme conditions remains an unresolved challenge, requiring advanced simulation environments, stress-testing methodologies, and formal verification techniques.

\subsubsection{Legal and Compliance Frameworks}
The dual nature of Neurosymbolic systems, merging the adaptive, data-driven insights of neural networks with the clear, rule-based logic of symbolic reasoning, complicates the clear attribution of accountability in incident scenarios. This complexity necessitates the establishment of legal guidelines that delineate responsibility among manufacturers, programmers, and operators, ensuring that each stakeholder is adequately accountable for AI-driven decisions. Moreover, given the global nature of AAM operations, harmonizing international standards becomes imperative to ensure that AI systems adhere to uniform safety and legal norms across diverse regulatory landscapes \cite{bengio2025international}. Achieving this will require coordinated efforts between legal experts, regulatory bodies, and technical developers to create transnational frameworks that not only standardize certification protocols but also provide mechanisms for continuous oversight and post-incident analysis.

\subsubsection{Data Governance and Privacy Policies}
The implementation of dynamic, adaptive policy frameworks is essential to ensure compliance with evolving data protection laws and international privacy standards. This approach requires continuous updates to data management practices, aligning with regulations like General Data Protection Regulation (GDPR) and California Consumer Privacy Act (CCPA) \cite{van2021data}. Transparency in data use is achieved through standardized protocols for data sharing among stakeholders, while sensitive information is protected using advanced encryption, anonymization, and access control measures. This focus on compliance and transparency not only builds trust among stakeholders but also creates a resilient ecosystem with meticulous documentation of data provenance and maintenance of audit trails for rigorous oversight \footnote{\url{https://www.oecd.org/content/dam/oecd/en/publications/reports/2024/06/ai-data-governance-and-privacy_2ac13a42/2476b1a4-en.pdf}}.

\section{Conclusion}
 Neurosymbolic AI holds significant promise for advancing AAM by delivering transparent, adaptable, and high-performance solutions tailored to meet the stringent demands of modern air transportation.  However, empirical insights from FAA and EASA initiatives expose pressing issues like scalability, data integration, cybersecurity, and transparent decision-making that must be rigorously addressed through standardized protocols and robust validation. Continued interdisciplinary research, along with sustained collaboration among academia, industry, and regulatory agencies, is essential to fully harness the potential of these hybrid approaches and to drive the next generation of safe, efficient, and sustainable air mobility solutions.

\appendix

\section*{Acknowledgments}
This material is based upon work supported by the NASA Aeronautics Research Mission Directorate (ARMD) University Leadership Initiative (ULI) under cooperative agreement number 80NSSC23M0059. This research was also partially supported by the U.S. National Science Foundation through Grant No. 2317117 and Grant No. 2309760.

\bibliographystyle{named}
\bibliography{ijcai25}

\begin{thebibliography}{}

\bibitem[\protect\citeauthoryear{Acharya \bgroup \em et al.\egroup }{2023}]{acharya2023neurosymbolic}
Kamal Acharya, Waleed Raza, Carlos Dourado, Alvaro Velasquez, and Houbing~Herbert Song.
\newblock Neurosymbolic reinforcement learning and planning: A survey.
\newblock {\em IEEE Transactions on Artificial Intelligence}, 2023.

\bibitem[\protect\citeauthoryear{Acharya \bgroup \em et al.\egroup }{2025}]{acharya2025neurosymbolic}
Kamal Acharya, Mehul Lad, Liang Sun, and Houbing Song.
\newblock Neurosymbolic ai for travel demand prediction: Integrating decision tree rules into neural networks.
\newblock {\em arXiv preprint arXiv:2502.01680}, 2025.

\bibitem[\protect\citeauthoryear{Barrera \bgroup \em et al.\egroup }{2022}]{barrera2022next}
Thomas~P Barrera, James~R Bond, Marty Bradley, Rob Gitzendanner, Eric~C Darcy, Michael Armstrong, and Chao-Yang Wang.
\newblock Next-generation aviation li-ion battery technologies—enabling electrified aircraft.
\newblock {\em The Electrochemical Society Interface}, 31(3):69, 2022.

\bibitem[\protect\citeauthoryear{Bengio \bgroup \em et al.\egroup }{2025}]{bengio2025international}
Yoshua Bengio, S{\"o}ren Mindermann, Daniel Privitera, Tamay Besiroglu, Rishi Bommasani, Stephen Casper, Yejin Choi, Philip Fox, Ben Garfinkel, Danielle Goldfarb, et~al.
\newblock International ai safety report.
\newblock {\em arXiv preprint arXiv:2501.17805}, 2025.

\bibitem[\protect\citeauthoryear{Bridgelall and Tolliver}{2024}]{bridgelall2024transforming}
Raj Bridgelall and Denver Tolliver.
\newblock Transforming healthcare delivery with advanced air mobility: A rural study with gis-based optimization.
\newblock {\em Sustainability}, 16(13):5709, 2024.

\bibitem[\protect\citeauthoryear{Cohen and Shaheen}{2024}]{cohen2024advanced}
Adam Cohen and Susan Shaheen.
\newblock Advanced air mobility: Opportunities, challenges, and research needsfor the state of california (2023-2030).
\newblock 2024.

\bibitem[\protect\citeauthoryear{Deniz \bgroup \em et al.\egroup }{2024}]{deniz2024reinforcement}
Sabrullah Deniz, Yufei Wu, Yang Shi, and Zhenbo Wang.
\newblock A reinforcement learning approach to vehicle coordination for structured advanced air mobility.
\newblock {\em Green Energy and Intelligent Transportation}, 3(2):100157, 2024.

\bibitem[\protect\citeauthoryear{Do \bgroup \em et al.\egroup }{2024}]{do2024cellular}
Hieu Do, Ramon~Delgado Pulgar, G{\'a}bor Fodor, and Zhiqiang Qi.
\newblock Cellular connectivity for advanced air mobility: Use cases and beamforming approaches.
\newblock {\em IEEE Communications Standards Magazine}, 8(1):65--71, 2024.

\bibitem[\protect\citeauthoryear{Dorn-Gomba \bgroup \em et al.\egroup }{2020}]{dorn2020power}
Lea Dorn-Gomba, John Ramoul, John Reimers, and Ali Emadi.
\newblock Power electronic converters in electric aircraft: Current status, challenges, and emerging technologies.
\newblock {\em IEEE Transactions on Transportation Electrification}, 6(4):1648--1664, 2020.

\bibitem[\protect\citeauthoryear{Dulia \bgroup \em et al.\egroup }{2021}]{dulia2021benefits}
Esrat~F Dulia, Mir~S Sabuj, and Syed~AM Shihab.
\newblock Benefits of advanced air mobility for society and environment: A case study of ohio.
\newblock {\em Applied Sciences}, 12(1):207, 2021.

\bibitem[\protect\citeauthoryear{Ellis \bgroup \em et al.\egroup }{2020}]{ellis2020time}
Kyle Ellis, John Koelling, Misty Davies, and Paul Krois.
\newblock In-time system-wide safety assurance (issa) concept of operations and design considerations for urban air mobility (uam).
\newblock 2020.

\bibitem[\protect\citeauthoryear{Ferrer \bgroup \em et al.\egroup }{2021}]{ferrer2021bias}
Xavier Ferrer, Tom Van~Nuenen, Jose~M Such, Mark Cot{\'e}, and Natalia Criado.
\newblock Bias and discrimination in ai: a cross-disciplinary perspective.
\newblock {\em IEEE Technology and Society Magazine}, 40(2):72--80, 2021.

\bibitem[\protect\citeauthoryear{Garcez and Lamb}{2023}]{garcez2023neurosymbolic}
Artur~d’Avila Garcez and Luis~C Lamb.
\newblock Neurosymbolic ai: The 3 rd wave.
\newblock {\em Artificial Intelligence Review}, 56(11):12387--12406, 2023.

\bibitem[\protect\citeauthoryear{Gilpin and Ilievski}{2021}]{gilpin2021neuro}
Leilani~H Gilpin and Filip Ilievski.
\newblock Neuro-symbolic reasoning in the traffic domain.
\newblock {\em J AI Res}, 15(3):123--145, 2021.

\bibitem[\protect\citeauthoryear{Goyal and Cohen}{2022}]{goyal2022advanced}
Rohit Goyal and Adam Cohen.
\newblock Advanced air mobility: Opportunities and challenges deploying evtols for air ambulance service.
\newblock {\em Applied Sciences}, 12(3):1183, 2022.

\bibitem[\protect\citeauthoryear{Goyal \bgroup \em et al.\egroup }{2021}]{goyal2021advanced}
Rohit Goyal, Colleen Reiche, Chris Fernando, and Adam Cohen.
\newblock Advanced air mobility: Demand analysis and market potential of the airport shuttle and air taxi markets.
\newblock {\em Sustainability}, 13(13):7421, 2021.

\bibitem[\protect\citeauthoryear{Grov \bgroup \em et al.\egroup }{2024}]{grov2024use}
Gudmund Grov, Jonas Halvorsen, Magnus~Wiik Eckhoff, Bj{\o}rn~Jervell Hansen, Martin Eian, and Vasileios Mavroeidis.
\newblock On the use of neurosymbolic ai for defending against cyber attacks.
\newblock In {\em International Conference on Neural-Symbolic Learning and Reasoning}, pages 119--140. Springer, 2024.

\bibitem[\protect\citeauthoryear{G{\"u}r \bgroup \em et al.\egroup }{2024}]{gur2024v2v}
G{\"u}rkan G{\"u}r, Kamesh Namuduri, Stefano Savazzi, Saba Al-Rubaye, Ashwin Ashok, Sven~G Bil{\'e}n, Ivan Petrunin, Marco Hernandez, Shane Nicoll, and Batool Dalloul.
\newblock V2v uas communications and use cases for advanced air mobility.
\newblock In {\em 2024 IEEE Conference on Standards for Communications and Networking (CSCN)}, pages 103--107. IEEE, 2024.

\bibitem[\protect\citeauthoryear{Hagos and Rawat}{2024}]{hagos2024neuro}
Desta~Haileselassie Hagos and Danda~B Rawat.
\newblock Neuro-symbolic ai for military applications.
\newblock {\em IEEE Transactions on Artificial Intelligence}, 2024.

\bibitem[\protect\citeauthoryear{Hill \bgroup \em et al.\egroup }{2020}]{hill2020uam}
Brian~P Hill, Dwight DeCarme, Matt Metcalfe, Christine Griffin, Sterling Wiggins, Chris Metts, Bill Bastedo, Michael~D Patterson, and Nancy~L Mendonca.
\newblock Uam vision concept of operations (conops) uam maturity level (uml) 4.
\newblock 2020.

\bibitem[\protect\citeauthoryear{Jalaian and Bastian}{2023}]{jalaian_neurosymbolic_2023}
Brian Jalaian and Nathaniel~D. Bastian.
\newblock Neurosymbolic ai in cybersecurity: Bridging pattern recognition and symbolic reasoning.
\newblock In {\em IEEE MILCOM 2023}, pages 268--273, October 2023.
\newblock ISSN: 2155-7586.

\bibitem[\protect\citeauthoryear{Jansen \bgroup \em et al.\egroup }{2020}]{jansen2020safe}
Nils Jansen, Bettina K{\"o}nighofer, Sebastian Junges, Alex Serban, and Roderick Bloem.
\newblock Safe reinforcement learning using probabilistic shields.
\newblock In {\em 31st International Conference on Concurrency Theory (CONCUR 2020)}. Schloss-Dagstuhl-Leibniz Zentrum f{\"u}r Informatik, 2020.

\bibitem[\protect\citeauthoryear{Johnson and Silva}{2022}]{johnson2022nasa}
W~Johnson and C~Silva.
\newblock Nasa concept vehicles and the engineering of advanced air mobility aircraft.
\newblock {\em The Aeronautical Journal}, 126(1295):59--91, 2022.

\bibitem[\protect\citeauthoryear{Karampinis \bgroup \em et al.\egroup }{2024}]{karampinis2024ensuring}
Vasileios Karampinis, Anastasios Arsenos, Orfeas Filippopoulos, Evangelos Petrongonas, Christos Skliros, Dimitrios Kollias, Stefanos Kollias, and Athanasios Voulodimos.
\newblock Ensuring uav safety: A vision-only and real-time framework for collision avoidance through object detection, tracking, and distance estimation.
\newblock {\em arXiv preprint arXiv:2405.06749}, 2024.

\bibitem[\protect\citeauthoryear{Kautz}{2022}]{kautz2022third}
Henry Kautz.
\newblock The third ai summer: Aaai robert s. engelmore memorial lecture.
\newblock {\em Ai magazine}, 43(1):105--125, 2022.

\bibitem[\protect\citeauthoryear{Khanmohamadi and Guerrieri}{2024}]{su16198336}
Masoud Khanmohamadi and Marco Guerrieri.
\newblock Advanced sensor technologies in cavs for traditional and smart road condition monitoring: A review.
\newblock {\em Sustainability}, 16(19), 2024.

\bibitem[\protect\citeauthoryear{Kiesewetter \bgroup \em et al.\egroup }{2023}]{KIESEWETTER2023100949}
Lukas Kiesewetter, Kazi~Hassan Shakib, Paramvir Singh, Mizanur Rahman, Bhupendra Khandelwal, Sudarshan Kumar, and Krishna Shah.
\newblock A holistic review of the current state of research on aircraft design concepts and consideration for advanced air mobility applications.
\newblock {\em Progress in Aerospace Sciences}, 142:100949, 2023.

\bibitem[\protect\citeauthoryear{Kimura \bgroup \em et al.\egroup }{2021}]{kimura2021reinforcement}
Daiki Kimura, Subhajit Chaudhury, Akifumi Wachi, Ryosuke Kohita, Asim Munawar, Michiaki Tatsubori, and Alexander Gray.
\newblock Reinforcement learning with external knowledge by using logical neural networks.
\newblock {\em arXiv preprint arXiv:2103.02363}, 2021.

\bibitem[\protect\citeauthoryear{Kohaut \bgroup \em et al.\egroup }{2024}]{kohaut2024probabilistic}
Simon Kohaut, Benedict Flade, Daniel Ochs, Devendra~Singh Dhami, Julian Eggert, and Kristian Kersting.
\newblock Probabilistic mission design in neuro-symbolic systems.
\newblock {\em arXiv preprint arXiv:2501.01439}, 2024.

\bibitem[\protect\citeauthoryear{Long \bgroup \em et al.\egroup }{2023}]{LONG2023102436}
Qi~Long, Jun Ma, Feifeng Jiang, and Christopher~John Webster.
\newblock Demand analysis in urban air mobility: A literature review.
\newblock {\em Journal of Air Transport Management}, 112:102436, 2023.

\bibitem[\protect\citeauthoryear{Mao \bgroup \em et al.\egroup }{2019}]{mao2019neuro}
Jiayuan Mao, Chuang Gan, Pushmeet Kohli, Joshua~B Tenenbaum, and Jiajun Wu.
\newblock The neuro-symbolic concept learner: Interpreting scenes, words, and sentences from natural supervision.
\newblock {\em arXiv preprint arXiv:1904.12584}, 2019.

\bibitem[\protect\citeauthoryear{Mur{\c{c}}a \bgroup \em et al.\egroup }{2024}]{murcca2024characterizing}
Mayara Cond{\'e}~Rocha Mur{\c{c}}a, Wallace Silva~Sant’Anna Souza, and Jo{\~a}o Vitor~Turchetti Ribeiro.
\newblock Characterizing performance tradeoffs from air traffic management services for advanced air mobility.
\newblock {\em Journal of Aerospace Information Systems}, pages 1--14, 2024.

\bibitem[\protect\citeauthoryear{Namuduri}{2023}]{namuduri2023digital}
Kamesh Namuduri.
\newblock Digital twin approach for integrated airspace management with applications to advanced air mobility services.
\newblock {\em IEEE Open Journal of Vehicular Technology}, 2023.

\bibitem[\protect\citeauthoryear{{National Academies of Sciences et al.}}{2018}]{national2018time}
{National Academies of Sciences et al.}
\newblock {\em In-time Aviation Safety Management: Challenges and Research for an Evolving Aviation System}.
\newblock National Academies Press, 2018.

\bibitem[\protect\citeauthoryear{Nguyen \bgroup \em et al.\egroup }{2022}]{nguyen2022artificial}
Van-Thai Nguyen, Phuc Do, Alexandre Vosin, and Benoit Iung.
\newblock Artificial-intelligence-based maintenance decision-making and optimization for multi-state component systems.
\newblock {\em Reliability Engineering \& System Safety}, 228:108757, 2022.

\bibitem[\protect\citeauthoryear{Novelli \bgroup \em et al.\egroup }{2024}]{novelli2024accountability}
Claudio Novelli, Mariarosaria Taddeo, and Luciano Floridi.
\newblock Accountability in artificial intelligence: what it is and how it works.
\newblock {\em Ai \& Society}, 39(4):1871--1882, 2024.

\bibitem[\protect\citeauthoryear{Petit and Shladover}{2014}]{petit2014potential}
Jonathan Petit and Steven~E Shladover.
\newblock Potential cyberattacks on automated vehicles.
\newblock {\em IEEE Transactions on Intelligent transportation systems}, 16(2):546--556, 2014.

\bibitem[\protect\citeauthoryear{Raghunatha \bgroup \em et al.\egroup }{2023}]{raghunatha2023addressing}
Aishwarya Raghunatha, Patrik Thollander, and Stephan Barthel.
\newblock Addressing the emergence of drones--a policy development framework for regional drone transportation systems.
\newblock {\em Transportation Research Interdisciplinary Perspectives}, 18:100795, 2023.

\bibitem[\protect\citeauthoryear{Rizzi and Scata~Jr}{2022}]{rizzi2022urban}
Stephen~A Rizzi and Donald~S Scata~Jr.
\newblock Urban air mobility community noise test planning.
\newblock In {\em 183rd Meeting of the Acoustical Society of America}, 2022.

\bibitem[\protect\citeauthoryear{Sharifi \bgroup \em et al.\egroup }{2023}]{sharifi2023towards}
Iman Sharifi, Mustafa Yildirim, and Saber Fallah.
\newblock Towards safe autonomous driving policies using a neuro-symbolic deep reinforcement learning approach.
\newblock {\em arXiv preprint arXiv:2307.01316}, 2023.

\bibitem[\protect\citeauthoryear{Silva \bgroup \em et al.\egroup }{2018}]{silva2018vtol}
Christopher Silva, Wayne~R Johnson, Eduardo Solis, Michael~D Patterson, and Kevin~R Antcliff.
\newblock Vtol urban air mobility concept vehicles for technology development.
\newblock In {\em 2018 Aviation Technology, Integration, and Operations Conference}, page 3847, 2018.

\bibitem[\protect\citeauthoryear{Siyaev \bgroup \em et al.\egroup }{2023}]{siyaev2023interaction}
Aziz Siyaev, Dilmurod Valiev, and Geun-Sik Jo.
\newblock Interaction with industrial digital twin using neuro-symbolic reasoning.
\newblock {\em Sensors}, 23(3):1729, 2023.

\bibitem[\protect\citeauthoryear{Van~Dalsem \bgroup \em et al.\egroup }{2021}]{van2021data}
William Van~Dalsem, Sandeep Shetye, Aditya~N Das, Kalmanje~S Krishnakumar, Sandy Lozito, Kenneth Freeman, Aaron Swank, Peter Shannon, and Luka Tomljenovic.
\newblock A data \& reasoning fabric to enable advanced air mobility.
\newblock In {\em AIAA Scitech 2021 Forum}, page 2033, 2021.

\bibitem[\protect\citeauthoryear{Wang \bgroup \em et al.\egroup }{2024}]{wang2024towards}
Wenguan Wang, Yi~Yang, and Fei Wu.
\newblock Towards data-and knowledge-driven ai: a survey on neuro-symbolic computing.
\newblock {\em IEEE Transactions on Pattern Analysis and Machine Intelligence}, 2024.

\bibitem[\protect\citeauthoryear{Yang \bgroup \em et al.\egroup }{2018}]{yang2018optimized}
Shunkun Yang, Chong Bian, Xing Li, Lin Tan, and Dongxiao Tang.
\newblock Optimized fault diagnosis based on fmea-style cbr and bn for embedded software system.
\newblock {\em The International Journal of Advanced Manufacturing Technology}, 94:3441--3453, 2018.

\bibitem[\protect\citeauthoryear{Yoo \bgroup \em et al.\egroup }{2022}]{yoo2022risk}
Jaeho Yoo, Yunseon Choe, and Soo-i Rim.
\newblock Risk perceptions using urban and advanced air mobility (uam/aam) by applying a mixed method approach.
\newblock {\em Sustainability}, 14(24):16338, 2022.

\bibitem[\protect\citeauthoryear{Yu \bgroup \em et al.\egroup }{2023}]{yu2023survey}
Dongran Yu, Bo~Yang, Dayou Liu, Hui Wang, and Shirui Pan.
\newblock A survey on neural-symbolic learning systems.
\newblock {\em Neural Networks}, 2023.

\end{thebibliography}

\end{document}